%% file: bare_jrnl_new_sample4.tex
\documentclass[10pt,journal,compsoc]{IEEEtran}
\usepackage{amsmath,amsfonts}
\usepackage{array}
\usepackage{longtable,tabularx,multicol}
\usepackage[caption=false,font=normalsize,labelfont=sf,textfont=sf]{subfig}
\usepackage{textcomp}
\usepackage{stfloats}
\usepackage{url}
\usepackage{verbatim}
\usepackage{graphicx}
\usepackage{cite}
\usepackage{bm}
\usepackage{amssymb}
\usepackage{amsmath}
\usepackage{makecell}
\usepackage{booktabs}
\usepackage{graphicx}
\usepackage{algorithmic} 
\usepackage{tikz} 
\usepackage[ruled,vlined]{algorithm2e} 
\usepackage[breaklinks,colorlinks,urlcolor=black]{hyperref}
\usepackage{colortbl,hhline}
\usepackage{multirow}
\usepackage{marvosym}
\usepackage{enumitem}
\usepackage{subcaption}
\usepackage{wrapfig}
\usepackage{xcolor}
\newcommand{\eg}{\emph{e.g.}}

\newcommand{\ours}{\textsc{PerceptRL}}

\newcommand{\Racc}{R_{\mathrm{acc}}}
\newcommand{\Rground}{R_{\text{ground}}}

\newcommand{\Rtotal}{R_{\text{total}}}

\newcommand{\target}{\texttt{\#target}}

\definecolor{mygray}{gray}{.92}

\begin{document}
\bstctlcite{perceptrl:BSTcontrol}

\title{\fontsize{22}{26}\selectfont
PerceptRL: Making Visual Perception Rewardable\\
in MLLMs from Answers to Interactions}
\author{Tongkun~Guan$^{1,2}$,~Zhibo~Yang$^{2,\dagger}$,~Mingkun~Yang$^{2}$,~Haocheng~Wang$^{1}$,\\
Wei~Shen$^{1}$,~Xiaokang~Yang$^{1}$,~\IEEEmembership{Fellow,~IEEE}%
\thanks{This work was supported by the National Natural Science Foundation of China under Grants 62322604 and 62576207. (Corresponding author: Wei Shen).}
\thanks{$^{1}$T. Guan, H. Wang, W. Shen, and X. Yang are with the MoE Key Laboratory of Artificial Intelligence, AI Institute, School of Computer Science, Shanghai Jiao Tong University, Shanghai 200240, China (e-mail: gtk0615@sjtu.edu.cn; wei.shen@sjtu.edu.cn; xkyang@sjtu.edu.cn).}
\thanks{$^{2}$T. Guan, Z. Yang, and M. Yang are with Qwen.}
\thanks{$^{\dagger}$Project leader: Zhibo Yang.}
}

\markboth{A SUBMISSION TO IEEE TRANSACTION ON PATTERN ANALYSIS AND MACHINE INTELLIGENCE}%
{Guan et al.: PerceptRL: Making Visual Perception Rewardable in MLLMs from Answers to Interactions}

\IEEEpubid{0000--0000/00\$00.00~\copyright~2021 IEEE}

\input{sec/0_abstract}
\input{sec/1_intro}
\input{sec/2_relat}
\input{sec/3_methodology}
\input{sec/4_experiment}

\input{sec/5_ablation}
\input{sec/6_conclusion}




 
\bibliographystyle{IEEEtran}
\bibliography{perceptrl_verified}

\newpage

 




\vfill

\end{document}

%% file: sec/0_abstract.tex
\IEEEtitleabstractindextext{
\begin{abstract}
Fine-grained visual perception remains a bottleneck during post-training of multimodal large language models (MLLMs), where models first perceive images and then reason over the extracted information, so perceptual errors can limit the gains from reasoning-focused optimization. Although extensive image-caption pre-training improves perception, the influence of this supervision can be diluted as later training stages change the data mixture and objectives. Continuing to optimize captions during post-training is a natural response, but their subjective, information-dense nature makes verifiable reward design difficult. To address this, we present PerceptRL to strengthen perception during post-training at two complementary levels. Perception RL uses a multi-agent pipeline to extract verified error items from dense image captions and compose the corresponding visual facts into rewardable QA pairs: chain QA links correlated errors, while multiple-choice QA combines unrelated errors. A specialized route targets high-error domains with tailored verification, while a general route addresses perceptual errors beyond these domains through a domain-agnostic pipeline. This error-driven construction provides fixed answer labels and rule-based rewards, making caption-derived perceptual supervision verifiable. Perception Agentic RL targets fine-grained and dense cases that single-turn Perception RL still fails to resolve, enabling multi-turn inspection with visual tools. We observed coordinate drift even in models trained to invoke tools: incorrect crops or annotations can distort later evidence and propagate errors. Coordinate Injection corrects spatial operations in cold-start trajectories, while a grounding-aware process reward supervises intermediate spatial fidelity alongside final-answer correctness. Training proceeds through shared cold-start SFT, single-turn Perception RL, and multi-turn Perception Agentic RL. Across seven benchmarks under a shared tool-use protocol, the full pipeline achieves 71.6\% average accuracy, exceeding the base model with the same tools and the strongest evaluated frontier VLM by 15.7 and 5.2 percentage points, respectively.
\end{abstract}

}

\maketitle

\IEEEdisplaynontitleabstractindextext

\IEEEpeerreviewmaketitle

%% file: sec/1_intro.tex
\IEEEraisesectionheading{\section{Introduction}\label{sec:intro}}
\begin{figure*}[!t]
\centering
\includegraphics[width=\textwidth]{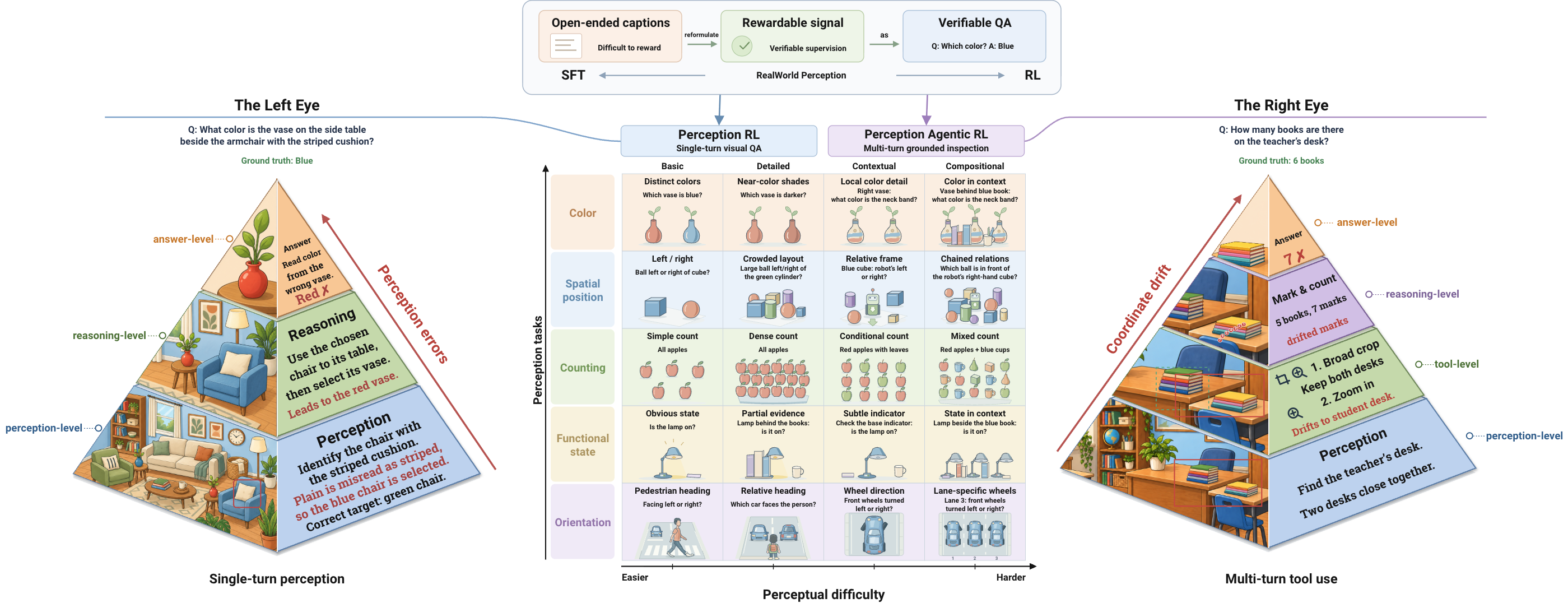}
\caption{\ours{}: making real-world perception rewardable, and training it at two levels.
\textbf{Center: How it becomes rewardable.} Open-ended captions contain useful perceptual information but are difficult to score with direct rule-based rewards. We extract and verify perceptual facts from these captions, then formulate QA pairs with fixed answer labels for rule-based training (top).
The pipeline organizes tasks from multiple perception families (\emph{e.g.,} color, spatial position, counting, functional state, orientation), using a perception tree to guide data construction (Section~\ref{sec:level1}).
The difficulty axis is what divides the two levels.
\textbf{Level~1: Perception RL.} A single look suffices here, so we train single-turn visual QA under a rule-based outcome reward.
The left panel shows what this must fix: an error entering at the perception level is inherited by an otherwise correct reasoning chain, so the answer is wrong for a purely perceptual reason.
\textbf{Level~2: Perception Agentic RL.} This tail, where objects are small, scenes are cluttered, or details require magnification, resists a single look, so the model inspects the image over multiple grounded turns with a code interpreter.
The right panel illustrates how a grounding error, where a predicted crop or annotation misses its target, can affect subsequent observations. Its propagation and accumulation across successive tool interactions constitute \emph{coordinate drift}. Coordinate Injection corrects these grounding errors during data construction, while a grounding-aware process reward supervises intermediate spatial fidelity (Section~\ref{sec:level2}).}
\label{fig:overview}
\end{figure*}

\IEEEPARstart{T}{he} progress of multimodal large language models (MLLMs) is fundamentally constrained by the quality of visual perception.
No matter how sophisticated a model's reasoning capabilities, if it fails at the perceptual level (\eg, miscounting objects, confusing colors, or misjudging spatial relationships), downstream reasoning may produce wrong answers~\cite{wu2026relationship,yang2026distractions}.
Fine-grained perception therefore remains an important concern for current MLLMs~\cite{qwen2.5vl,llava_onevision,gpt4o,xu2025lvlmehub}.
CodePercept~\cite{codepercept} finds that scaling perception is more effective than scaling reasoning in its analysis of STEM visual reasoning.
A perceptual error can compromise an otherwise valid reasoning chain (Figure~\ref{fig:overview}, left).

Image--caption data provide an important source of visual supervision during the pre-training stage~\cite{zhang2024vlmsurvey,radford2021clip,li2023blip2}.
However, production MLLMs undergo multiple stages, including ViT pre-training, vision--language alignment, instruction tuning, reinforcement learning, and on-policy distillation~\cite{qwen2.5vl,qwen3vl}.
As the data mixtures and training objectives change, the influence of perception-focused captions used in the pre-training stage can be diluted by later optimization.

A natural response is to continue optimizing captions during post-training, with reinforcement learning (RL) providing feedback on model-generated outputs.
For example, in mathematical reasoning, final answers can often be checked against reference solutions, giving RL objective feedback on generated solutions without requiring annotations for every reasoning step~\cite{deepseek_r1,openai_o1}.
Direct caption optimization, however, encounters two further difficulties.
\textbf{Subjectivity}: factually correct captions can differ in granularity, emphasis, style, and length, making the desired optimization target difficult to specify.
\textbf{Unrewardability}: captions are information-dense and open-ended, bundling numerous visual claims into one description. Constructing a verifiable reward requires deciding which claims to assess and how to combine their correctness into a caption-level score.

Solver-based methods~\cite{caprl} reward captions according to how well a vision-free LLM solver can answer visual questions using those captions.
This is an indirect perception signal: a high reward indicates that the caption provides information useful to the sampled questions and that the solver can exploit it, but does not establish that the caption captures the image accurately and comprehensively.
Visual details not required by those questions may remain unexamined.

Code-based methods~\cite{codepercept} use image-to-code-to-image reproduction to supervise visual perception, using code executability and reconstruction fidelity as reward signals.
However, this formulation is developed for structured STEM images that can be rendered programmatically; its generalization to diverse real-world images remains unestablished.

These three issues raise a central question: \emph{how can we reinforce visual perception during post-training through objective optimization targets and verifiable rewards, so that the learned skills generalize across diverse perception tasks?}
We present \ours{}, a framework that makes perceptual supervision explicit at two complementary levels: single-turn answers and multi-turn visual interactions (Figure~\ref{fig:overview}).

\noindent\textbf{Level 1: Perception RL.} Our key insight is that the base model's own perceptual errors can guide the construction of targeted, verifiable training data.
We design a multi-agent pipeline that first elicits dense captions from the base model, audits their visual claims, and cross-verifies detected errors against independent vision models.
The critical step is \emph{error composition}: we turn verified visual facts associated with these errors into \emph{rewardable QA pairs}, combining unrelated errors into multiple-choice questions and linking correlated errors into multi-step visual identification chains.
With fixed answer labels supported by cross-verification and human review, these questions provide rule-based outcome rewards, making caption-derived perceptual supervision explicit and verifiable.
Guided by a perception tree, specialized pipelines focus on selected high-error domains, while an \emph{error-agnostic} general pipeline discovers and composes failures beyond those domains.
Both routes follow this \emph{error-driven} principle, concentrating single-turn Perception RL on the model's demonstrated perceptual weaknesses.

\noindent\textbf{Level 2: Perception Agentic RL.} For fine-grained, complex, and dense perceptual errors, where objects are small, scenes are cluttered, or details require magnification, single-turn answering often fails.
Tool use through a code interpreter can substantially help by enabling the model to inspect image regions over multiple turns.
Prior work has explored learning visual tool-use policies and integrating code execution into reasoning~\cite{deepeyes,deepeyesv2,revpt,toolsrl}.
In multi-turn visual inspection, however, choosing an appropriate tool or operation does not by itself ensure that the predicted crop or annotation lands on the intended target.
A \emph{grounding error} occurs when a crop box or annotation point misses its intended target.
An incorrect crop may omit relevant evidence, while a misplaced annotation can mislead subsequent localization or counting.
If later turns depend on these observations, the initial error may propagate and compound across interactions.
We refer to this process as \emph{coordinate drift} (Figure~\ref{fig:overview}, right).
This motivates a complementary focus on intermediate spatial fidelity alongside final-answer correctness.
To this end, Coordinate Injection (CI) constructs cold-start trajectories with corrected spatial coordinates, while a grounding-aware process reward supervises the spatial fidelity of intermediate operations during multi-turn RL.

Training follows three sequential stages: a single shared cold-start SFT stage on a mixture of perception QA pairs and CI-corrected trajectories, followed by single-turn Perception RL and then multi-turn Perception Agentic RL. Each stage starts from the preceding checkpoint; there is no additional SFT stage between the two RL stages.

Our contributions are as follows:
\begin{itemize}[nosep,leftmargin=1.5em]
    \item We propose Perception RL to address the difficulty of defining verifiable rewards for open-ended image captions. It converts verified perceptual facts associated with the model's caption errors into QA pairs with fixed answer labels, enabling single-turn post-training of visual perception with rule-based outcome rewards. Across eight perception benchmarks, this stage raises average accuracy from $61.9\%$ after cold-start SFT to $69.8\%$ (Table~\ref{tab:main_results}).
    \item We propose Perception Agentic RL for fine-grained and dense perception requiring multi-turn inspection. To mitigate coordinate drift, Coordinate Injection corrects spatial operations in cold-start trajectories, while a grounding-aware process reward supervises intermediate grounding alongside answer correctness. Under the shared tool-use protocol, the full pipeline achieves $71.6\%$ average accuracy across seven benchmarks, outperforming the base model with the same tools and the strongest evaluated frontier VLM by $15.7$ and $5.2$ percentage points, respectively (Table~\ref{tab:agentic_cross_model}).
\end{itemize}

%% file: sec/2_relat.tex
\section{Related Work}
\label{sec:related}

\subsection{Fine-Grained Visual Perception}
Fine-grained visual perception requires recovering local details and their spatial structure~\cite{wang2024ptp,tang2024transcp}.
Grounding DINO~\cite{grounding_dino} combines language-guided feature fusion and grounded pre-training to detect objects specified by category names or referring expressions.
SAM~\cite{kirillov2023sam} learns promptable segmentation from large-scale mask supervision, while Florence-2~\cite{xiao2024florence2} unifies captioning, detection, grounding, and segmentation through a common sequence-to-sequence formulation.

Text-centric methods improve detection through feature refinement and synthetic-to-real pre-training~\cite{guan2022industrial,guan2024bridging}, and recognition through glyph alignment and character-level distillation~\cite{guan2023siga,guan2023self,guan2025ccdplus}.
Structured document perception further incorporates symbol positions for mathematical expressions~\cite{guan2024posformer}, token-level supervision for dense text~\cite{guan2025token}, and joint question answering and mask generation~\cite{wang2025marten}.

LReSeg~\cite{wen2026exploring} addresses reasoning segmentation with a small language model, using text-conditioned register tokens and a unified visual encoder for reasoning and mask decoding.
Wen et al.~\cite{wen2026efficient} reduce the cost of MLLM-based segmentation through delayed mask-token insertion and proxy-guided visual-token updates, preserving spatial details for dense prediction.
These studies emphasize explicit alignment between visual representations and spatially localized outputs.

In the medical domain, Hulu-Med~\cite{jiang2025hulu} unifies text, 2D images, 3D volumes, and videos in a generalist VLM, using medical-aware visual token reduction to improve computational efficiency.
\ours{} investigates how to provide verifiable perception supervision across domains and spatial supervision for successive tool interactions during reinforcement learning.

\subsection{Reinforcement Learning for VLMs}

\noindent\textbf{Reasoning-focused.} Reinforcement learning with verifiable rewards (RLVR) has emerged as a promising paradigm for enhancing reasoning in large language models~\cite{deepseek_r1,openai_o1}.
Its extension to vision-language models has attracted growing attention~\cite{yindiscriminative, sun2026see, chen2026videostarbench, chen-etal-2025-datasets}.
Vision-R1~\cite{vision_r1} constructs a 200K multimodal CoT dataset for cold-start initialization and employs GRPO~\cite{grpo} with progressive thinking suppression training.
VLM-R1~\cite{vlm_r1} applies rule-based RL to vision-language tasks with a focus on training stability and reward hacking mitigation.
R1-VL~\cite{r1vl} introduces step-wise rewards to improve logical consistency in multimodal reasoning.
G1~\cite{g1} discovers that perception and reasoning abilities can mutually bootstrap each other during RL training through a perception-enhanced cold start on visual games.
AVAR~\cite{avar} proposes the Visual Attention Score (VAS) to quantify visual token attention and designs attention-guided objectives to address lazy attention localization.
Despite these advances, the above methods primarily target reasoning capabilities, improving the logical process from visual evidence to conclusions, while leaving the perceptual accuracy of the visual evidence itself largely unaddressed.

\noindent\textbf{Perception-focused.} Perception-R1~\cite{perception_r1} demonstrates via McNemar's test that standard RLVR does not improve perception and proposes a visual perception reward for explicit perceptual supervision.
PAPO~\cite{papo} introduces an implicit perception loss based on KL divergence between distributions conditioned on original versus corrupted visual inputs, achieving significant perception error reduction without external annotations.
V-Triune~\cite{vtriune} proposes a Dynamic IoU reward for detection and grounding tasks, avoiding reward ambiguity under loose thresholds and reward sparsity under strict ones.
CodePercept~\cite{codepercept} establishes code as a perceptual medium for STEM visual understanding, showing that scaling perception consistently outperforms scaling reasoning.
CapRL~\cite{caprl} redefines caption quality through downstream utility, rewarding captions that enable a vision-free LLM to answer visual questions, and demonstrates that RL-trained captioners can match much larger models.

\subsection{Thinking with Python}
A growing line of work explores using code execution as a reasoning tool for VLMs~\cite{gupta2023visprog,suris2023vipergpt}.
DeepEyes~\cite{deepeyes} learns to ``think with images'' via RL, using the model's own grounding capability to dynamically crop and re-examine image regions during reasoning.
DeepEyesV2~\cite{deepeyesv2} extends this to an agentic setting that interleaves code execution and web search within a single reasoning trajectory, employing a cold-start SFT stage followed by RL to establish robust tool-use patterns.
R1-Code-Interpreter~\cite{r1ci} trains LLMs to autonomously generate code during step-by-step reasoning via multi-stage curriculum RL.
ReVPT~\cite{revpt} enhances VLMs' ability to reason about and use visual tools through RL.
ToolsRL~\cite{toolsrl} proposes a two-stage curriculum that decouples tool mastery from answer optimization.
Spatial accuracy is a complementary concern in such tool-use settings, because later interactions may depend on the crops and annotations produced in earlier turns.

\ours{} differs from prior work in two fundamental ways.
First, at the \emph{data level}, we automatically mine the base model's own perceptual errors and compose them into verifiable QA pairs, producing training signal that is both error-agnostic and scalable.
Second, at the \emph{supervision level}, we combine Coordinate Injection for constructing spatially corrected training trajectories with a grounding-aware process reward that evaluates intermediate spatial fidelity during RL.

%% file: sec/3_methodology.tex
\section{Method}
\label{sec:method}

\subsection{Overview}
\label{sec:overview}

\ours{} operates at two complementary levels (Figure~\ref{fig:overview}).
Level~1 (Perception RL) systematically probes the model's perceptual weaknesses, converts the identified errors into verifiable QA pairs, and trains single-turn perception with a rule-based outcome reward.
However, its single-turn RL does not optimize iterative visual inspection, and small objects, densely packed scenes, or details requiring magnification can remain difficult to resolve from the full image alone.
This motivates Level~2 (Perception Agentic RL), which trains multi-turn visual inspection through a code interpreter for cropping, resizing, annotation, and measurement.
Such interactions require accurate spatial grounding: an incorrect crop or annotation can distort the evidence used in later turns and cause coordinate drift.

A single multi-agent pipeline underlies both levels, turning the base model's own failures into training signal (Section~\ref{sec:data}): verifiable single-turn QA pairs for Level~1 and coordinate-corrected multi-turn agentic trajectories for Level~2.
On top of this data, the model is trained in three sequential stages, each initialized from the previous checkpoint (Figure~\ref{fig:overview}):
(1)~a single \emph{cold-start SFT} on a mixture of the single-turn perception QA and the multi-turn CI-corrected trajectories, which provides training examples for single-turn perception QA and multi-turn tool use with CI-corrected coordinates;
(2)~\emph{Level-1 Perception RL}, single-turn RL from the cold-start checkpoint that trains on perception QA with a rule-based outcome reward; and
(3)~\emph{Level-2 Perception Agentic RL}, multi-turn RL that continues from the Level-1 checkpoint and additionally supervises the spatial fidelity of every intermediate step with a grounding-aware process reward.
The two levels describe the single-turn and multi-turn task settings; the three stages describe the training sequence.

\subsection{Perception RL}
\label{sec:level1}

Current post-training is dominated by reasoning RL, which has substantially improved reasoning ability precisely because reasoning tasks admit abundant rewardable signals.
Perception, in contrast, has seen far less progress in the post-training stage: its natural supervision, captions, is subjective and open-ended, and thus difficult to formulate as a verifiable reward.
Perceptual errors can undermine subsequent reasoning, motivating explicit supervision of visual evidence~\cite{perception_r1}.
This motivates Level~1: to provide verifiable supervision across multiple perception task families during post-training.

\begin{figure*}[t]
\begin{center}
\includegraphics[width=\textwidth]{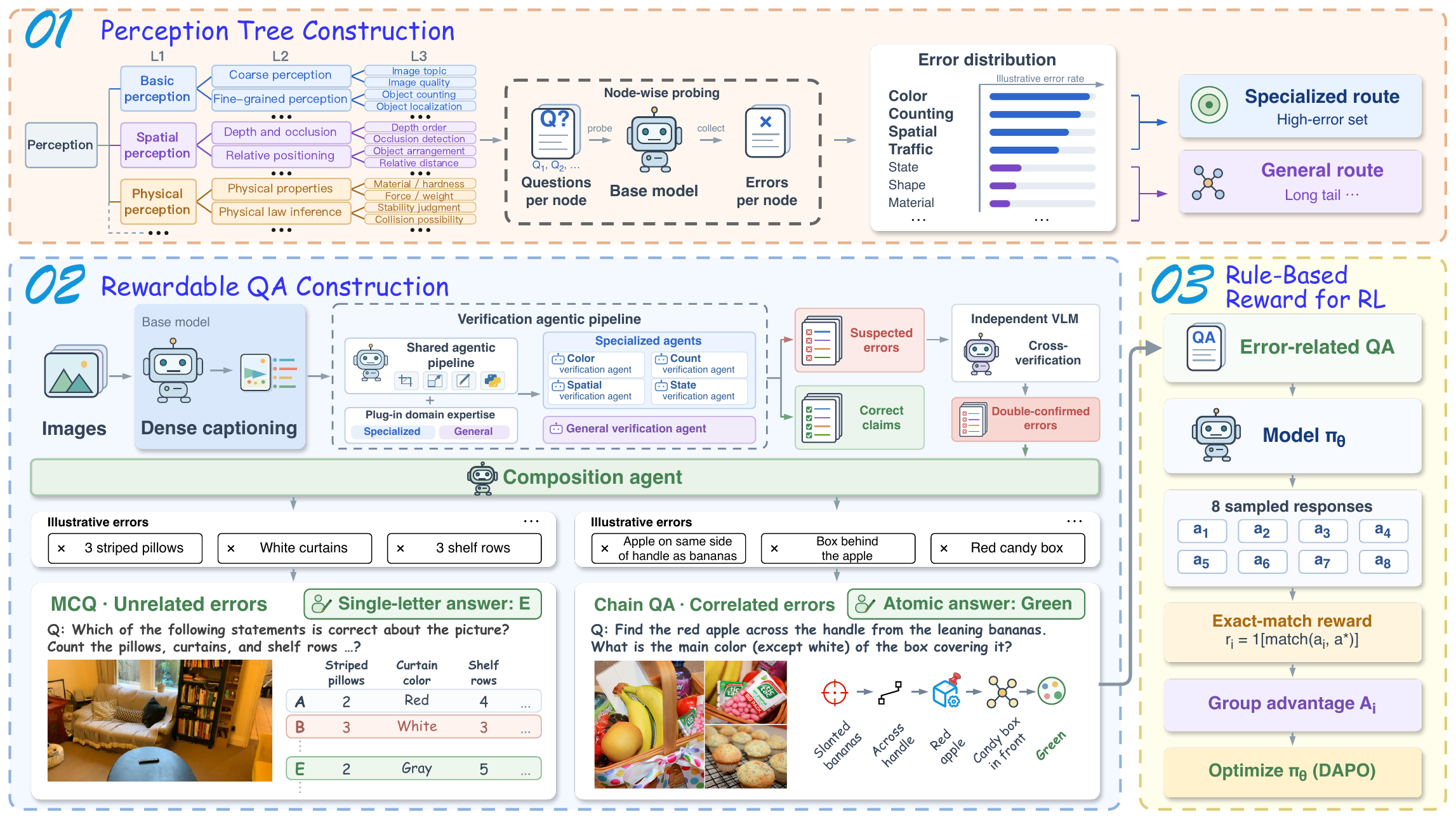}
\end{center}
\caption{\textbf{Perception RL: from perceptual errors to rewardable QA.}
\textbf{(01)~Perception tree construction:} A hierarchical capability map and model probing reveal perceptual weaknesses, guiding specialized and general data-construction routes. Appendix~\ref{app:perception_tree} expands the displayed tree with additional skills and associated failure modes.
\textbf{(02)~Rewardable QA construction:} A multi-agent pipeline mines and verifies errors in the base model's captions, then composes them into questions with unambiguous answers. MCQs combine unrelated errors, while chain QA connects correlated errors.
\textbf{(03)~Rule-based reward for RL:} The resulting QA pairs provide exact-match rewards for training the model's visual perception.
The error distribution and example errors are illustrative.}
\label{fig:perception_rl}
\end{figure*}

Training across multiple perception skills calls for structured task organization: perception includes a broad range of fine-grained abilities, such as counting, color, orientation, and depth.
We therefore organize Level~1 as a three-step pipeline (Fig.~\ref{fig:perception_rl}).
\textbf{(i)}~We construct a three-level perception tree to organize fine-grained capabilities and guide model probing and data sampling. Two complementary routes support this construction: a specialized route for selected high-error domains and a general route for additional error types (Section~\ref{sec:tree}).
\textbf{(ii)}~Along both routes, a multi-agent pipeline turns the base model's own perceptual errors into deterministic, verifiable QA (Section~\ref{sec:data}).
\textbf{(iii)}~We train on this data with rule-based reward RL (Section~\ref{sec:perception_rl}).

\subsubsection{Perception Tree Construction}
\label{sec:tree}
Perception is not a single scalar ability: a model may read colors reliably yet systematically miscount, or localize objects well yet confuse facing directions.
A structured decomposition helps identify perceptual weaknesses and organize data construction across related skills.
We therefore build a three-level perception tree that progressively organizes broad aspects of visual understanding into related capability groups and concrete perceptual tasks (Figure~\ref{fig:perception_rl}).
The hierarchy covers visual attributes, spatial relationships, and physical understanding, connecting high-level capability organization to the specific errors that guide training-data construction.
The figure illustrates representative branches of this organization, while Table~\ref{tab:perception_tree} in Appendix~\ref{app:perception_tree} provides a more detailed capability map and associated failure modes.

With the tree in place, we probe the base model on sampled tasks associated with its capabilities and examine where its answers disagree with the reference labels.
The observed error patterns guide data allocation and the selection of task families for specialized construction, while the general route covers additional errors beyond these selected domains.
We use the following two construction routes:
\begin{itemize}[nosep,leftmargin=1.5em]
    \item \textbf{Specialized route.} For high-error nodes whose answers can be made cleanly and objectively verifiable, we build dedicated per-domain pipelines, currently for counting, color, spatial, and state, each with domain-specific verification (\eg, pixel-level color sampling, per-object box counting) and question templates.
    \item \textbf{General route.} A single domain-agnostic pipeline handles errors beyond the selected specialized domains.
It discovers errors in the base model's image descriptions and composes them into questions, providing a flexible way to sample additional task types.
\end{itemize}
Both routes share the same error-driven construction principle (Section~\ref{sec:data}); they differ only in whether verification and composition are specialized to one node or applied generically.
The tree provides an organizational framework for data construction; empirical gains are assessed on the benchmark tasks reported in Section~\ref{sec:exp}.

\subsubsection{Rewardable QA Construction}
\label{sec:data}
The central insight shared by both routes is that perceptual errors are more informative than correct perceptions: rather than training on what the model already knows, we systematically discover what it gets wrong and build targeted data around those failures.
This decomposes data construction into two sub-problems: (i)~reliably detecting the model's genuine perceptual errors, and (ii)~composing those errors into QA that is both challenging (it targets an actual weakness) and verifiable (it has an unambiguous correct answer).
We surface genuine errors through a three-stage multi-agent protocol.
\textbf{(1)~Dense captioning}: the base model produces an exhaustive multi-dimensional caption over the categories above.
\textbf{(2)~Agent-based verification}: a VLM audits the caption claim by claim (global checks on the full image, local checks on crops with quantitative measurement, and a scan for omissions), emitting each error as a structured JSON record.
\textbf{(3)~Cross-verification}: a second, independently chosen VLM re-checks every error, and only those confirmed by both are retained, filtering out single-model false positives.
\textbf{(4)~Error composition}: rather than filling fixed templates, a dedicated composition agent turns confirmed errors into QA pairs end-to-end, choosing which errors to target, phrasing them into one coherent question, and pairing the correct answer with crafted distractors under the quality gates below. Because it reasons about each error configuration rather than slotting values into a schema, it adapts question format, difficulty, and distractors to the errors an image exhibits, and chooses between two strategies by how the errors relate:

\noindent\textbf{MCQ format (unrelated errors).} When an image contains multiple independent perceptual errors (\eg, a color error, a counting error, and a spatial error that concern different objects), we compose them into a multiple-choice question where each option (A--H) combines multiple visual claims.
Exactly one option has all claims correct; the distractors each contain one or more claims that correspond to verified errors.
This format produces questions with unambiguous single-letter answers, ideal for rule-based reward.

\noindent\textbf{Chain QA format (correlated errors).} When errors are spatially or semantically related (\eg, multiple spatial errors involving the same group of objects), we compose them into a multi-step visual identification chain: an initial candidate set $\to$ discriminating filter $\to$ intermediate anchor $\to$ further discrimination $\to$ unique target $\to$ atomic answer.
Each step in the chain requires a distinct perceptual capability, and the answer is only reachable by correctly executing all steps.

Both formats yield deterministic, verifiable answers, and because every question is grown from a genuine error the base model made, learning concentrates on its weaknesses rather than rehearsing what it already perceives correctly.
This pipeline runs in the two routes of Section~\ref{sec:tree}: the \textbf{general route} composes errors sampled from different capability pools into one hard question, while the \textbf{specialized route} applies the same skeleton with domain-specific verification and templates to the highest-error, cleanly-verifiable nodes (Appendix~\ref{app:pipelines}).
Both are governed by quality gates (capability coupling across $\geq$3 skills, distractor resistance, anti-shortcut tests, explicit reference frames), and a final round of manual inspection screens the constructed QA pairs.
Human review serves as the final quality-control stage for the constructed questions and their answer labels, following automated cross-verification.
The rule-based reward used during RL then checks model outputs against these reviewed labels, separating label validation during data construction from reward computation during training.

\subsubsection{Rule-Based Reward for RL}
\label{sec:perception_rl}
The verified QA pairs provide fixed answer labels, allowing perception learning to use a rule-based outcome reward.
For a model answer $a$ and its reference label $a^*$, the reward is
\begin{equation}
\Racc(a,a^*)=
\begin{cases}
1, & \text{match}(a,a^*),\\
0, & \text{otherwise}.
\end{cases}
\label{eq:reward_outcome}
\end{equation}
The target is an atomic value: a single letter (A--H) for MCQ, or a color, number, or direction for chain QA and counting.
The matching rule checks the final answer against this label; incorrect or unparseable answers receive zero reward.
Label verification and human review take place during data construction, so reward computation during RL requires no caption judge.

We optimize the policy with Decoupled Clip and Dynamic Sampling Policy Optimization (DAPO)~\cite{dapo}.
For each image--question input $x=(I,q)$, we sample $G$ responses $\{o_i\}_{i=1}^{G}$ from the rollout policy $\pi_{\theta_{\mathrm{old}}}(\cdot\mid x)$.
Let $a_i$ be the final answer in response $o_i$ and $R_i=\Racc(a_i,a^*)$ its outcome reward.
The group-normalized advantage is
\begin{equation}
\begin{aligned}
\bar{R} &= \frac{1}{G}\sum_{j=1}^{G}R_j,\\
\sigma_R &= \operatorname{std}(R_1,\ldots,R_G),\\
\widehat{A}_{i,t} &= \frac{R_i-\bar{R}}{\sigma_R},
\qquad 1\leq t\leq |o_i|.
\end{aligned}
\label{eq:dapo_advantage}
\end{equation}
Here, $t$ indexes generated tokens and $|o_i|$ is the response length.
Because Level~1 uses an outcome reward, every token in a response receives the same advantage, determined by its answer's correctness relative to the other responses in the group.

For token $o_{i,t}$ with preceding response tokens $o_{i,<t}$, the probability ratio and its clipped counterpart are
\begin{equation}
\begin{aligned}
u_{i,t}(\theta)
&=\frac{\pi_\theta(o_{i,t}\mid x,o_{i,<t})}
{\pi_{\theta_{\mathrm{old}}}(o_{i,t}\mid x,o_{i,<t})},\\
\widetilde{u}_{i,t}(\theta)
&=\operatorname{clip}\bigl(u_{i,t}(\theta),
1-\varepsilon_{\mathrm{low}},1+\varepsilon_{\mathrm{high}}\bigr).
\end{aligned}
\label{eq:dapo_ratio}
\end{equation}
The lower and upper clipping ranges are decoupled; DAPO's Clip-Higher strategy uses a larger upper range to allow greater probability increases for positively advantaged tokens.
Writing $N=\sum_{i=1}^{G}|o_i|$, the policy maximizes the token-averaged objective
\begin{equation}
\begin{aligned}
\mathcal{J}_{\mathrm{DAPO}}(\theta)
&=\mathbb{E}_{\mathcal{B}_{\mathrm{dyn}}}\Biggl[
\frac{1}{N}\sum_{i=1}^{G}\sum_{t=1}^{|o_i|}\\
&\quad\min\Bigl(
u_{i,t}(\theta)\widehat{A}_{i,t},\,
\widetilde{u}_{i,t}(\theta)\widehat{A}_{i,t}
\Bigr)\Biggr].
\end{aligned}
\label{eq:dapo_objective}
\end{equation}
The expectation is over image--question inputs and their sampled response groups retained by dynamic sampling, denoted by $\mathcal{B}_{\mathrm{dyn}}$.
For the binary outcome reward, a group is retained only when
\begin{equation}
0<\sum_{i=1}^{G}R_i<G.
\label{eq:dapo_dynamic_sampling}
\end{equation}
Groups whose responses are all correct or all incorrect are discarded, and sampling continues until the training batch is filled with informative groups.
This ensures nonzero reward variance for Equation~\ref{eq:dapo_advantage}.
The normalization by $N$ in Equation~\ref{eq:dapo_objective} averages over generated tokens rather than assigning equal weight to responses of different lengths.

\subsection{Perception Agentic RL}
\label{sec:level2}

\begin{figure*}[t]
\begin{center}
\includegraphics[width=\textwidth]{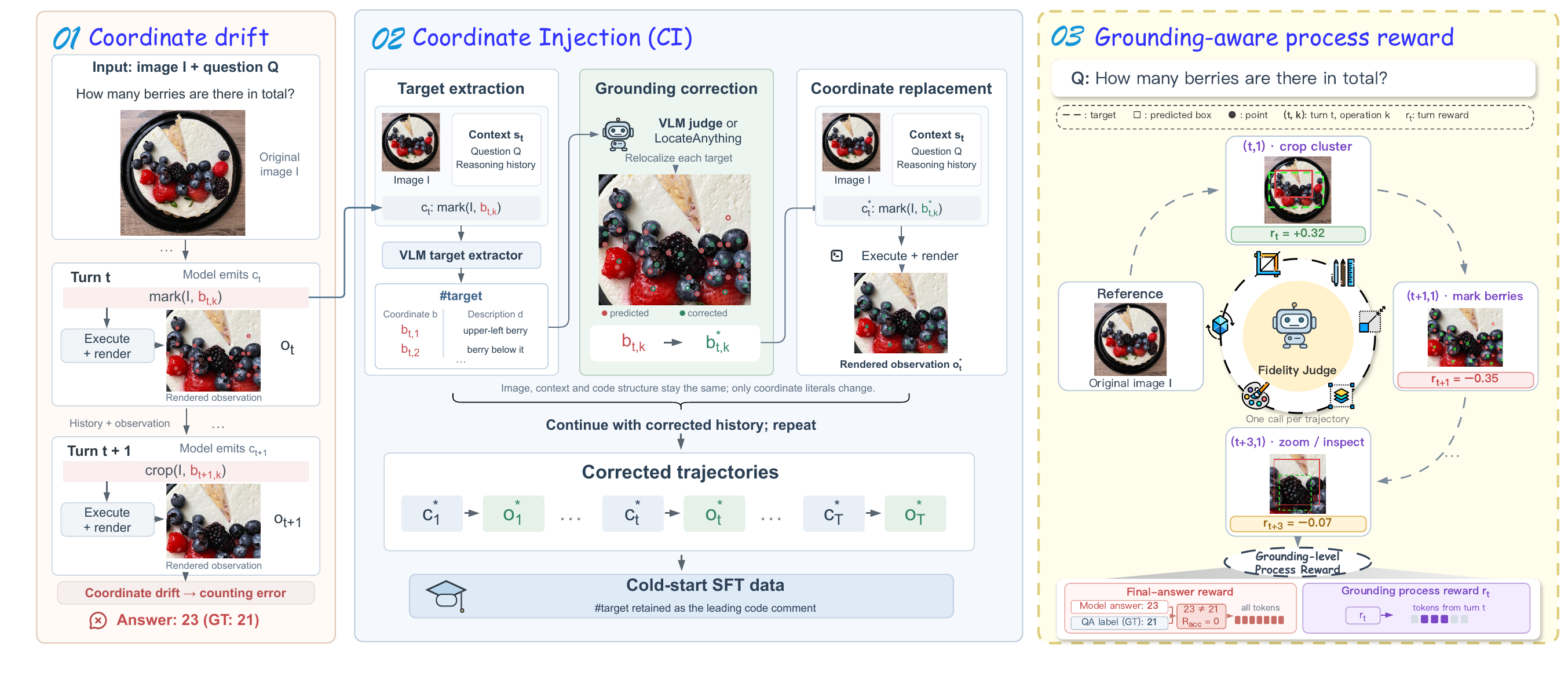}
\end{center}
\caption{\textbf{Perception Agentic RL: correcting and supervising spatial grounding during multi-turn visual inspection.}
\textbf{(01)~Coordinate drift.} A misplaced annotation changes the visual evidence available to later turns. Subsequent spatial operations can inherit this error, leading to repeated or missed counts.
\textbf{(02)~Coordinate Injection (CI).} During trajectory construction, CI identifies the intended target of each spatial operation, corrects its coordinates using a VLM judge or the LocateAnything detector, and re-executes the code before the next turn. The corrected trajectories provide cold-start SFT data.
\textbf{(03)~Grounding-aware process reward.} During RL, a fidelity judge evaluates the predicted boxes and points against their intended targets in the original image. The resulting per-turn rewards complement the final-answer reward and supervise intermediate spatial accuracy (Equation~\ref{eq:reward_turn_total}).}
\label{fig:agentic}
\end{figure*}

Level~2 targets difficult perceptual queries involving small objects, cluttered scenes, or subtle color and state differences, for which a single view may not provide sufficient evidence.
The model therefore uses a code interpreter to crop, resize, annotate, and measure the image over multiple turns before answering.
A \emph{grounding error} occurs when a crop box or annotation point misses its intended target.
If later turns rely on the resulting observation, this error can alter subsequent localization and counting decisions; we refer to the propagation and accumulation of such spatial misalignment as \emph{coordinate drift}.
For example, misplaced marks in dense counting can cause later turns to revisit counted objects or omit unmarked ones (Fig.~\ref{fig:agentic}, panel~01).
Final-answer correctness alone does not identify which intermediate spatial operations were accurate.
We address this supervision gap through two components: Coordinate Injection constructs trajectories with corrected spatial coordinates for cold-start SFT, and a grounding-aware process reward scores intermediate spatial fidelity during Level-2 RL.

\subsubsection{Coordinate Injection}
\label{sec:ci}

Given an image $I$ and a question $Q$, the agent alternates between generating a code block and observing its execution result.
Let $s_t$ denote the accumulated image-question prompt, reasoning context, and interaction history available before turn $t$.
The multi-turn interaction is
\begin{equation}
\begin{aligned}
s_1 &= (I,Q),\\
c_t &\sim \pi_\theta(\,\cdot\mid s_t),\\
o_t &= \mathrm{Exec}(I,c_t),\\
s_{t+1} &= s_t\oplus(c_t,o_t),
\end{aligned}
\label{eq:agent_interaction}
\end{equation}
where $\pi_\theta$ is the model policy, $\mathrm{Exec}$ is the code interpreter, $o_t$ is the resulting visual observation, and $\oplus$ appends the generated code and observation to the context.
After $T$ tool turns, the model generates its final answer from $s_{T+1}$.
The code-observation sequence is
\begin{equation}
\tau=(c_1,o_1,\ldots,c_T,o_T).
\label{eq:agent_trajectory}
\end{equation}
Each turn operates on the original image $I$; spatial operations, coordinate correction, and fidelity scoring use the coordinate system of $I$.

Equation~\ref{eq:agent_interaction} shows how a grounding error can enter the history: an off-target crop or annotation changes $o_t$, which becomes part of the context used to generate the next code block.
To reduce this source of coordinate drift in the training trajectories, Coordinate Injection (CI) corrects the spatial coordinates and re-executes the code before appending the observation.
Thus, subsequent generation uses the corrected observation rather than the output of the inaccurate operation.
Writing the corrected context as $s_t^*$, with $s_1^*=(I,Q)$, CI applies the following three steps to each turn that contains spatial operations (Fig.~\ref{fig:agentic}, panel~02).

\noindent\textit{1) Target extraction.}
For the current code block $c_t$, a VLM reads $I$, the context $s_t^*$, and the code, and produces
\begin{equation}
\mathcal{T}_t
=\mathrm{Extract}(I,s_t^*,c_t)
=\{(b_{t,k},d_{t,k})\}_{k=1}^{K_t}.
\label{eq:ci_target_extraction}
\end{equation}
Here, $K_t$ is the number of spatial operations in $c_t$, $b_{t,k}$ is the coordinate used by operation $k$, and $d_{t,k}$ describes its intended target.
The extraction prompt requires $b_{t,k}$ to be copied from the corresponding operation in the code.
The \target{} declaration serializes $\mathcal{T}_t$ as a JSON list with one record per operation.
For example, the following schema shows a box operation followed by a point operation; the underbraces identify the mathematical quantity represented by each field:
\begingroup\small
\[
\left[
\begin{aligned}
&\texttt{\{"bbox":}\ 
  \underbrace{\texttt{[x1,y1,x2,y2]}}_{b_{t,1}},\\[-1pt]
&\quad\texttt{"desc":}\ 
  \underbrace{\texttt{"upper-left berry"}}_{d_{t,1}}
  \texttt{\}},\\[3pt]
&\texttt{\{"point":}\ 
  \underbrace{\texttt{[x,y]}}_{b_{t,2}},\\[-1pt]
&\quad\texttt{"desc":}\ 
  \underbrace{\texttt{"lower berry"}}_{d_{t,2}}
  \texttt{\}},\ \ldots
\end{aligned}
\right].
\]
\endgroup
The subscript $t$ identifies the turn, and $k$ identifies the operation within that turn.
The \texttt{bbox} and \texttt{point} fields are two representations of $b_{t,k}$: a box has coordinates $[x_1,y_1,x_2,y_2]$, whereas a point has coordinates $[x,y]$.
The \texttt{desc} field contains $d_{t,k}$; the symbolic subscripts and underbraces are explanatory annotations and are not part of the JSON.

\noindent\textit{2) Grounding correction.}
For each extracted operation, a grounding backend uses the intended-target description to obtain a corrected reference coordinate:
\begin{equation}
b_{t,k}^*=\mathrm{Ground}(b_{t,k},d_{t,k},I).
\label{eq:ci_grounding}
\end{equation}
By default, a \emph{VLM judge} examines the predicted box or point on a coordinate-grid overlay of $I$ and adjusts its coordinates using $d_{t,k}$.
This backend handles most domains, including subtle attributes or functional states that are difficult to express as detection queries.
For domains with many precisely localizable targets, such as dense counting, we instead use an \emph{expert detector} (\eg, LocateAnything~\cite{locateanything}), which localizes the target from $I$ and $d_{t,k}$ without using the predicted coordinate.
The selected backend provides $b_{t,k}^*$ for the subsequent code correction.

\noindent\textit{3) Code rewriting and re-execution.}
CI replaces the coordinates associated with each operation, executes the corrected block, and appends its result to the corrected context:
\begin{equation}
\begin{aligned}
c_t^* &= \mathrm{Replace}
  \bigl(c_t,\{b_{t,k}\mapsto b_{t,k}^*\}_{k=1}^{K_t}\bigr),\\
o_t^* &= \mathrm{Exec}(I,c_t^*),\\
s_{t+1}^* &= s_t^*\oplus(c_t^*,o_t^*).
\end{aligned}
\label{eq:ci_reexecution}
\end{equation}
The model generates the next code block from $s_{t+1}^*$, and CI repeats the same procedure if that block contains spatial operations.
For a turn with no spatial operation, the code is executed and appended without coordinate replacement.
After all $T$ turns, this procedure yields
\begin{equation}
\tau^*=(c_1^*,o_1^*,\ldots,c_T^*,o_T^*),
\label{eq:ci_trajectory}
\end{equation}
where each grounding code block uses the corrected coordinates and each observation is obtained by executing its corresponding block.
We also update the \target{} records to contain $(b_{t,k}^*,d_{t,k})$ and retain them as a comment on the first line of each grounding code block.
These declarations couple an operation's coordinates to its intended-target description and can be parsed without analyzing the plotting code.
The resulting trajectories are included in the shared cold-start SFT data, teaching the model both multi-turn tool use and the declaration format later used for grounding rewards.

\subsubsection{Grounding-Aware Reward for RL}
\label{sec:rl}

The CI-corrected SFT trajectories teach the model to place a \target{} declaration at the beginning of each grounding code block.
During Level-2 RL, the model follows the interaction in Equation~\ref{eq:agent_interaction}, and each declaration is required to record the coordinates used by its corresponding spatial operation together with a description of the intended target.
The reward pipeline parses these records, renders the declared boxes or points on $I$, and evaluates whether they land on the described targets.
This provides a grounding signal for individual turns in addition to final-answer correctness.

As in CI, turn $t$ contains $K_t$ spatial operations, each indexed by $(t,k)$ and represented by $(b_{t,k},d_{t,k})$.
A non-grounding turn has $K_t=0$.
The set of turns eligible for grounding rewards is
\begin{equation}
\mathcal{G}=\{t\in\{1,\ldots,T\}:K_t>0\}.
\label{eq:grounding_turns}
\end{equation}
For each operation, the scorer assesses whether the declared box or point lands on the target named by $d_{t,k}$.
To avoid a separate judge call for each operation, all spatial operations in a trajectory are scored together using one composite image.

For operation $(t,k)$, we overlay $b_{t,k}$ on the original image and crop a padded window around it.
The resulting patch enlarges the annotation while retaining local context and is labeled with its index $(t,k)$ and description $d_{t,k}$.
We tile these patches with the unannotated original image as a global reference:
\begin{equation}
\begin{aligned}
P_{t,k}&=\mathrm{Patch}(I,b_{t,k},d_{t,k}),\\
M&=\mathrm{Tile}
 \bigl(I,\{P_{t,k}\}_{t\in\mathcal{G},\,1\leq k\leq K_t}\bigr).
\end{aligned}
\label{eq:grounding_mosaic}
\end{equation}
The fidelity judge reads $M$ in one call and returns a score $g_{t,k}\in[0,1]$ for each indexed patch.
A score of $1$ indicates a box that tightly encloses the described target or a point placed precisely on it; a score of $0$ indicates an off-target annotation.
Scores are returned as a JSON list keyed by the operation index, for example:
\begin{center}
\small\ttfamily
[\{"id":"t3\_k1","score":0.82\},\\
\{"id":"t3\_k2","score":0.15\}, \ldots]
\end{center}
The example values illustrate the output format.
Matching each returned index to its declaration recovers the score for the corresponding operation.

Each grounding turn is summarized by its least accurate operation:
\begin{equation}
G_t=\min_{1\leq k\leq K_t}g_{t,k},
\label{eq:reward_turn}
\end{equation}
so that correct operations cannot mask an off-target box or point in the same turn.
For a trajectory with at least one grounding turn, its mean grounding fidelity is
\begin{equation}
\bar{G}=\frac{1}{|\mathcal{G}|}\sum_{t\in\mathcal{G}}G_t.
\label{eq:gbar}
\end{equation}
The experimental metric averages $\bar{G}$ over trajectories containing at least one grounding turn.

For reward computation, we center the grounding score at $\tfrac12$:
\begin{equation}
\Rground(t)=G_t-\tfrac12,\qquad t\in\mathcal{G}.
\label{eq:grounding_centered}
\end{equation}
Scores above this threshold contribute positively and scores below it contribute negatively.
Each turn also receives the tool-use shaping term $\rho_t$, and the combined reward is clipped:
\begin{equation}
r_t=
\begin{cases}
\operatorname{clip}\bigl(\Rground(t)+\rho_t,-1,+1\bigr),
 &t\in\mathcal{G},\\[4pt]
\operatorname{clip}\bigl(\rho_t,-1,+1\bigr),
 &t\notin\mathcal{G}.
\end{cases}
\label{eq:reward_turn_total}
\end{equation}
Thus, non-grounding turns receive shaping rewards without a spatial-fidelity score.
The shaping term combines code-format validity, penalties for redundant display calls, runtime errors and cross-turn code duplication, and a bonus for repairing the preceding turn's runtime error.

\begin{table*}[t]
\caption{Main results on eight perception benchmarks, scored by accuracy ($\%$) under each benchmark's official protocol, comparing \ours{} against proprietary and open frontier VLMs. \textbf{Avg.} averages over the eight reported benchmarks; best per column in \textbf{bold}.}
\label{tab:main_results}
\centering
\resizebox{\textwidth}{!}{
\begin{tabular}{lccccccccc}
\toprule
\textbf{Method} & \textbf{RealWorldQA} & \textbf{MMStar} & \textbf{BLINK} & \textbf{TreeBench} & \textbf{ERQA} & \textbf{RefSpatial} & \textbf{CountQA} & \textbf{RealWorldCounting} & \textbf{Avg.} \\ 
\midrule
\multicolumn{10}{l}{\textit{Frontier VLMs}} \\
Qwen3.5-Plus~\cite{qwen_visual_models} & 82.2 & 82.2 & 73.4 & 62.2 & 62.7 & 69.7 & 71.7 & 35.4 & 67.4 \\ 
Gemini-3-Flash~\cite{gemini_api_models} & 84.6 & 81.5 & \textbf{76.3} & 58.3 & \textbf{65.2} & 63.4 & 70.4 & 34.6 & 66.8 \\ 
Doubao-Seed-2.0-mini~\cite{seed20} & 82.2 & 77.5 & 73.8 & 56.3 & 52.2 & 63.4 & 70.2 & 37.9 & 64.2 \\ 
Qwen3.5-Flash~\cite{qwen_visual_models} & 72.8 & 74.9 & 66.3 & 58.8 & 61.3 & 57.9 & 67.8 & 22.0 & 60.2 \\ 
Kimi-K2.6~\cite{kimi_k26} & 80.0 & 79.2 & 74.4 & 55.6 & 48.8 & 43.3 & 43.2 & 12.9 & 54.7 \\ 
Kimi-K2.5~\cite{kimi_k25} & 80.9 & 78.5 & 75.9 & 52.6 & 51.2 & 39.5 & 46.5 & 13.4 & 54.8 \\ 
Claude-Opus-4.8~\cite{claude_opus48} & 77.1 & 75.4 & 70.2 & 58.3 & 51.5 & 41.9 & 34.2 & 17.8 & 53.3 \\ 
Grok-4.5~\cite{grok45} & 74.5 & 75.0 & 73.9 & 59.5 & 61.0 & 12.3 & 45.4 & 19.0 & 52.6 \\ 
Gemini-2.5-Pro~\cite{gemini_api_models} & 78.2 & 77.9 & 67.0 & 54.8 & 58.5 & 27.3 & 40.5 & 19.2 & 52.9 \\ 
Claude-Opus-4.7~\cite{claude_opus47} & 77.6 & 77.1 & 70.3 & 57.3 & 51.5 & 37.2 & 33.4 & 13.4 & 52.2 \\ 
DeepSeek-V4-Flash-Vision~\cite{deepseek_v4_vision} & 77.1 & 72.7 & 65.1 & 58.0 & 58.2 & 20.6 & 49.3 & 11.2 & 51.5 \\ 
Grok-4.3~\cite{grok43} & 79.3 & 74.0 & 69.2 & 53.1 & 48.0 & 16.3 & 35.9 & 14.9 & 48.8 \\ 
Claude-Opus-4.6~\cite{claude_opus46} & 75.4 & 74.3 & 68.8 & 53.3 & 46.0 & 5.8 & 17.3 & 13.8 & 44.3 \\ 
\midrule
\multicolumn{10}{l}{\ours{} \textit{(on Qwen3.6-Flash, 35A3)}} \\
Base (Qwen3.6-Flash)~\cite{qwen36_35b_a3b} & 82.9 & 78.7 & 68.8 & 57.8 & 58.2 & 56.3 & 66.8 & 24.3 & 61.7 \\ 
$+$ SFT cold-start & 83.3 & 77.9 & 68.6 & 56.8 & 53.8 & 67.7 & 57.9 & 29.3 & 61.9 \\ 
$+$ Perception RL (Level 1) & \textbf{85.2} & \textbf{82.6} & 75.1 & \textbf{64.7} & 62.8 & \textbf{73.8} & \textbf{71.9} & \textbf{42.1} & \textbf{69.8} \\ 
\bottomrule
\end{tabular}
}
\end{table*}

\subsubsection{Token-Level Credit Assignment}
\label{sec:token_credit}
Both terms are assigned at the token level: the answer-accuracy reward is broadcast to \emph{every} token of the trajectory, whereas each turn's reward $r_t$ reaches only the tokens $\mathcal{I}_t$ produced in that turn (the $\mathcal{I}_t$ partition the trajectory).
The reward placed on token $i$ is
\begin{equation}
\begin{aligned}
\Rtotal(\tau)_i
&= \underbrace{\Racc(a, a^*)}_{\text{answer accuracy (all tokens)}}\\
&\quad+\underbrace{\sum_{t=1}^{T} r_t\,\mathbf{1}[\,i\in\mathcal{I}_t\,]}_{\text{intermediate-turn supervision}},
\end{aligned}
\label{eq:reward_total}
\end{equation}
where $a$ is the final answer and $\Racc(a, a^*)$ is the Level-1 exact-match reward of Equation~\ref{eq:reward_outcome}, so each token receives the outcome reward plus the reward of the turn that produced it.
The two terms supply credit at complementary granularities.
The outcome term $\Racc$ is constant across all tokens of a trajectory, so it provides only \emph{across-trajectory} credit: under the group-normalized advantage it separates a better rollout from a worse one, but is blind to \emph{which} turn inside a rollout earned that outcome.
The per-turn term $r_t$ instead varies from turn to turn, providing \emph{within-trajectory} credit: a well-grounded turn is advantaged over a poorly grounded turn in the \emph{same} trajectory, even when both share the same final answer.
Grounding fidelity is invisible to the across-trajectory term alone, so it is precisely this within-trajectory signal that lets RL reinforce accurate intermediate grounding rather than leave it unconstrained.

%% file: sec/4_experiment.tex
\section{Experiments}
\label{sec:exp}

\subsection{Experimental Setup}
\label{sec:setup}

\noindent\textbf{Base model and training.}
We build \ours{} on Qwen3.6-Flash~\cite{qwen36_35b_a3b} as the base VLM, using the 35A3 mixture-of-experts variant ($\sim$35B total / $\sim$3B active parameters), and train it in three stages, each initialized from the previous stage's checkpoint:
(i)~a \emph{cold-start SFT} stage on 57,678 training examples comprising single-turn perception QA and CI-corrected multi-turn agentic trajectories from our pipeline, which jointly teaches the single-turn perception-answer format and the multi-turn perception-with-grounding format together with a drift-free multi-turn reasoning pattern;
(ii)~\emph{Perception RL} (Level~1), single-turn RL over 36,445 perception QA queries spanning multiple domains (Table~\ref{tab:data_stats}), scored by a rule-based outcome reward;
and (iii)~\emph{Perception Agentic RL} (Level~2), multi-turn RL over the 19,810-query Agentic QA subset, scored by the grounding-aware process reward.
In both RL stages the model receives only queries and generates its own rollouts, which the reward then scores; no pre-collected trajectory is used as a supervision target.
Both RL stages use DAPO~\cite{dapo}, and training is run on 64$\times$A100 80GB.
The two stages differ in their optimization budget: Level~1 (Perception RL) uses a learning rate of $1\times10^{-6}$, a batch size of 128, a DAPO group size of $G=8$, a maximum sequence length of 32K tokens, and 1 epoch, whereas Level~2 (Agentic RL) uses a learning rate of $5\times10^{-7}$, a batch size of 64, a group size of $G=8$, a maximum sequence length of 128K tokens, and 1 epoch.
During Level-2 rollouts the model drives a sandboxed Python code interpreter (crop, resize, annotate, and pixel/box measurement) for up to $T=64$ turns per trajectory.

\noindent\textbf{From single-turn to agentic QA.}
The two QA pools in Table~\ref{tab:data_stats} are constructed sequentially.
We first train the Level-1 model with Perception RL on the full Single-turn QA pool of 36,445 pairs.
After Level-1 training, we generate eight rollouts from the trained model for every query in this same pool and compute its empirical accuracy as $c_i/8$, where $c_i$ is the number of correct responses.
We retain queries satisfying $c_i/8<5/8$, equivalently at most four correct responses out of eight, to form the 19,810-pair Agentic QA pool used for Level-2 RL.
This step selects existing QA pairs without adding new ones, so Agentic QA is a strict subset of Single-turn QA; their sizes are not additive.

\noindent\textbf{Benchmarks.}
We evaluate on eight perception benchmarks, grouped by the capability they stress: general real-world VQA (RealWorldQA~\cite{realworldqa}, MMStar~\cite{mmstar}), multi-image and fine-grained grounded perception (BLINK~\cite{blink}, TreeBench~\cite{treebench}), embodied and spatial perception (ERQA~\cite{erqa}, which targets embodied reasoning, and RefSpatial-Bench~\cite{refspatial} for referring spatial grounding), and object counting (CountQA~\cite{countqa} and real-world RealWorldCounting).
None of these suites is used when constructing our training data, so they measure generalization rather than in-distribution fit.

\begin{table*}[t]
\caption{Comparison of multi-turn tool-use capabilities across frontier VLMs.
Cold-start SFT denotes the shared SFT checkpoint before Level-1 and Level-2 RL; its single-turn scores are taken from Table~\ref{tab:main_results}.
Accuracy is reported in $\%$. Avg is the mean over the seven benchmarks shown. Best per column in \textbf{bold}.}
\label{tab:agentic_cross_model}
\centering
\resizebox{\textwidth}{!}{
\begin{tabular}{lccccccc|c}
\toprule
\textbf{Model (agentic)} & \textbf{RealWorldQA} & \textbf{MMStar} & \textbf{BLINK} & \textbf{TreeBench} & \textbf{ERQA} & \textbf{CountQA} & \textbf{RealWorldCounting} & \textbf{Avg} \\ 
\midrule
\multicolumn{9}{l}{\textit{Frontier VLMs $+$ code interpreter}} \\
Gemini-3-Flash~\cite{gemini_api_models} & 84.4 & \textbf{82.8} & \textbf{76.9} & 62.2 & \textbf{71.8} & 53.7 & 33.2 & 66.4 \\ 
Doubao-Seed-2.0-mini~\cite{seed20} & 78.3 & 77.7 & 73.1 & 54.6 & 54.0 & 65.8 & 33.7 & 62.5 \\ 
Qwen3.5-Flash~\cite{qwen_visual_models} & 74.6 & 77.9 & 44.6 & 56.3 & 54.0 & 44.1 & 17.7 & 52.7 \\ 
Kimi-K2.6~\cite{kimi_k26} & 79.1 & 78.5 & 73.7 & 54.8 & 62.3 & 54.1 & 25.0 & 61.1 \\ 
Claude-Opus-4.8~\cite{claude_opus48} & 80.7 & 79.9 & 76.6 & 59.0 & 61.3 & 53.1 & 29.2 & 62.8 \\ 
Grok-4.5~\cite{grok45} & 85.0 & 81.4 & 73.2 & 64.0 & 64.0 & 54.3 & 23.9 & 63.7 \\ 
DeepSeek-V4-Flash-Vision~\cite{deepseek_v4_vision} & 79.6 & 79.3 & 75.4 & 62.5 & 63.7 & 58.4 & 28.2 & 63.9 \\ 
\midrule
\multicolumn{9}{l}{\textit{Our models $+$ code interpreter}} \\
Baseline~\cite{qwen36_35b_a3b} & 82.4 & 78.7 & 49.8 & 56.5 & 55.0 & 49.7 & 19.4 & 55.9 \\ 
Cold-start SFT & 83.3 & 77.9 & 68.6 & 56.8 & 53.8 & 57.9 & 29.3 & 61.1 \\ 
Perception Agentic RL (\textbf{Ours}) & \textbf{87.1} & 79.1 & 76.4 & \textbf{68.6} & 62.2 & \textbf{73.9} & \textbf{53.9} & \textbf{71.6} \\ 
\bottomrule
\end{tabular}
}
\vspace{4pt}
\end{table*}

\noindent\textbf{RefSpatial evaluation protocol.}
RefSpatial-Bench is evaluated using the official point-in-mask criterion after converting each model's predicted point to image-space $[x,y]$ coordinates.
Gemini predictions use normalized $[y,x]$ coordinates in the range $[0,1000]$; we reorder the axes and scale each coordinate by the corresponding image dimension.
Claude-Opus-4.6/4.7/4.8 predictions are interpreted directly as pixel coordinates.
For Grok-4.3 and Kimi-K2.5/K2.6, points in $[0,1]$ are interpreted as fractional image coordinates, and otherwise as $[0,1000]$ normalized coordinates; both representations are mapped to pixels before evaluation.
Table~\ref{tab:main_results} reports the resulting accuracies.
The remaining models use the benchmark's standard $[0,1000]$ $[x,y]$ convention.

\begin{table}[t]
\caption{Perception training QA across domains.
\textbf{Single-turn QA} is the full pool used for Perception RL (Level~1).
\textbf{Agentic QA} is a strict subset selected from this pool after Level-1 training: the trained model generates eight rollouts per query, and queries with empirical accuracy $<5/8$ are retained for Perception Agentic RL (Level~2).}
\label{tab:data_stats}
\centering
\resizebox{\columnwidth}{!}{
\begin{tabular}{llcc}
\toprule
\textbf{Route} & \textbf{Domain} & \textbf{Single-turn QA} & \textbf{Agentic QA} \\
\midrule
\multirow{4}{*}{Specialized} & Counting & 14,109 & 9,783 \\
 & Color & 3,710 & 377 \\
 & Spatial & 3,965 & 1,150 \\
 & State & 3,331 & 2,483 \\
\midrule
General & General Perception & 11,330 & 6,017 \\
\midrule
\multicolumn{2}{l}{\textbf{Total}} & \textbf{36,445} & \textbf{19,810} \\
\bottomrule
\end{tabular}
}
\end{table}

\subsection{Main Results}
\label{sec:main_results}

Table~\ref{tab:main_results} compares frontier VLMs with the successive training stages of \ours{} on the Qwen3.6-Flash (35A3) backbone.
The frontier models use single-turn inference, while the final Level-2 model uses multi-turn tool-augmented inference; Table~\ref{tab:agentic_cross_model} compares models under a shared tool interface.
The shared SFT cold-start initializes the perception-answer and grounded tool-use formats, obtaining $61.9\%$ mean accuracy compared with $61.7\%$ for the base model.
Perception RL then improves every benchmark relative to this checkpoint and raises mean accuracy to $69.8\%$, a gain of $7.9$ percentage points over the SFT model and $8.1$ points over the base model.
These improvements are achieved in the single-turn setting, before introducing Level-2 tool interactions.

The largest gains over the SFT checkpoint are on CountQA ($57.9\%$ to $71.9\%$), RealWorldCounting ($29.3\%$ to $42.1\%$), and ERQA ($53.8\%$ to $62.8\%$), corresponding to improvements of $14.0$, $12.8$, and $9.0$ percentage points.
The improvements also extend to fine-grained and spatial perception: TreeBench increases by $7.9$ points, BLINK by $6.5$ points, and RefSpatial by $6.1$ points.
The Level-1 checkpoint attains $73.8\%$ on RefSpatial, the highest accuracy among the models in Table~\ref{tab:main_results}.
Gains on RealWorldQA and MMStar further show that the improvement extends beyond counting and localization.

The final Level-2 model reaches $71.6\%$ mean accuracy under the shared tool-use protocol and achieves the best reported results on RealWorldQA, TreeBench, and RealWorldCounting in Table~\ref{tab:agentic_cross_model}.
RealWorldCounting reaches $53.9\%$, providing the largest improvement over Level~1.
Compared with Level~1, scores are lower on MMStar, ERQA, and RefSpatial.

\noindent\textbf{Multi-turn tool-use capabilities.}
Table~\ref{tab:agentic_cross_model} compares frontier VLMs with the base model and \ours{} on seven perception benchmarks.
The frontier VLMs, the base model, and \ours{} run in the same sandboxed code-interpreter loop (crop/resize/annotate/measure), capped at $T=64$ turns and a 128K-token budget, with the same tool prompt and official per-benchmark scoring; only the model varies.

Under the shared protocol, \ours{} achieves the highest mean accuracy of $71.6\%$, exceeding the strongest frontier baseline, Gemini-3-Flash, by $5.2$ percentage points.
It leads on RealWorldQA, TreeBench, CountQA, and RealWorldCounting, with its largest advantage on RealWorldCounting.
Performance remains task-dependent: Gemini-3-Flash leads on MMStar, BLINK, and ERQA.

Under the same tool interface, the base model attains $55.9\%$ mean accuracy.
The full training pipeline improves the reported mean accuracy by $15.7$ percentage points relative to this baseline, with the tool interface held fixed.

%% file: sec/5_ablation.tex
\section{Ablation Studies}
\label{sec:ablations}
\subsection{Coverage of the Two Construction Routes}
\label{sec:ablation_routes}
Level~1 builds data along two routes (Section~\ref{sec:tree}): a \emph{specialized} route with per-domain pipelines for the high-error nodes (counting, color, spatial, state) and a \emph{general} domain-agnostic route for the long tail.
To assess their respective contributions, we compare three Level-1 RL variants sharing the same SFT cold-start, RL recipe, and total query budget, differing only in which route supplies the data: specialized-only, general-only, and both.

As shown in Table~\ref{tab:routes_ablation}, the specialized route outperforms the general route on TreeBench, ERQA, RefSpatial, CountQA, and RealWorldCounting, with advantages of 6.7 and 6.2 percentage points on the two counting benchmarks, respectively.
The general route achieves higher accuracy on RealWorldQA, MMStar, and BLINK, indicating a complementary benefit from broader domain coverage.
Combining both routes yields the highest accuracy on all eight benchmarks, including 71.9\% on CountQA and 42.1\% on RealWorldCounting. These results support the complementary contributions of domain-specific construction and general coverage under a matched data budget.

\begin{table*}[t]
\caption{Coverage of the two Level-1 construction routes on the Qwen3.6-Flash (35A3) backbone.
All rows share the same SFT cold-start and RL recipe under a \emph{matched total query budget}, differing only in which route supplies the RL data; \emph{Both} is our full Perception RL. Best per column in \textbf{bold}.}
\label{tab:routes_ablation}
\centering
\resizebox{\textwidth}{!}{
\begin{tabular}{lcccccccc}
\toprule
\textbf{Method} & \textbf{RealWorldQA} & \textbf{MMStar} & \textbf{BLINK} & \textbf{TreeBench} & \textbf{ERQA} & \textbf{RefSpatial} & \textbf{CountQA} & \textbf{RealWorldCounting} \\
\midrule
SFT cold-start & 83.3 & 77.9 & 68.6 & 56.8 & 53.8 & 67.7 & 57.9 & 29.3 \\
$+$ Specialized route only & 83.6 & 78.9 & 72.0 & 63.2 & 61.0 & 72.4 & 70.2 & 40.0 \\
$+$ General route only & 84.8 & 81.7 & 73.6 & 59.5 & 57.2 & 69.6 & 63.5 & 33.8 \\
$+$ Both routes (\textbf{Ours}) & \textbf{85.2} & \textbf{82.6} & \textbf{75.1} & \textbf{64.7} & \textbf{62.8} & \textbf{73.8} & \textbf{71.9} & \textbf{42.1} \\
\bottomrule
\end{tabular}
}
\end{table*}

\begin{table*}[t]
\caption{Effect of inference-time coordinate correction on final-answer accuracy using Gemini-3.5-Flash (no training).
\textbf{$+$ with Python} acts on \emph{uncorrected} coordinates; \textbf{Ours} applies Coordinate-Injection correction at inference.
Parentheses show gains over the baseline; best per column in \textbf{bold}.}
\label{tab:ci_ablation}
\centering
\resizebox{\textwidth}{!}{
\begin{tabular}{lcccc}
\toprule
\textbf{Setting} & \textbf{RealWorldQA} & \textbf{TreeBench} & \textbf{PerceptionBench} & \textbf{RealWorldCounting} \\ 
\midrule
Gemini-3.5-Flash (baseline) & 85.4 & 60.5 & 69.4 & 33.0 \\ 
$+$ with Python (uncorrected coordinates) & 86.0 (+0.6) & 67.2 (+6.7) & 70.0 (+0.6) & 37.4 (+4.4) \\ 
$+$ with Python (\textbf{Ours}) & \textbf{87.8} (+2.4) & \textbf{68.9} (+8.4) & \textbf{74.4} (+5.0) & \textbf{45.8} (+12.8) \\ 
\bottomrule
\end{tabular}
}
\end{table*}

\begin{table*}[t]
\caption{Grounding fidelity $\bar{G}$ during Level-2 RL with outcome-only or grounding-aware process reward. Higher is better.}
\label{tab:grounding_curve}
\centering
\resizebox{\textwidth}{!}{
\begin{tabular}{lccccccccccc}
\toprule
\textbf{Step} & 0 & 20 & 40 & 60 & 80 & 100 & 120 & 140 & 160 & 180 & 200 \\
\midrule
Outcome-only ($\bar{G}$) & 0.55 & 0.57 & 0.63 & 0.59 & 0.66 & 0.58 & 0.64 & 0.60 & 0.66 & 0.59 & 0.63 \\
$+$ Grounding-aware process reward (\textbf{Ours}, $\bar{G}$) & 0.55 & 0.61 & 0.66 & 0.70 & 0.73 & 0.76 & 0.73 & 0.80 & 0.81 & 0.84 & 0.81 \\
\bottomrule
\end{tabular}
}
\end{table*}

\subsection{Comparison with Specialized Tool-Use Models}
\label{sec:ablation_specialized_tool_use}
Table~\ref{tab:specialized_tool_use} compares specialized visual tool-use models on V*~\cite{vstar}, HR-Bench~\cite{hrbench}, MMStar, and TreeBench.
The published V*, HR-Bench, and TreeBench scores are taken from the DeepEyesV2 study~\cite{deepeyesv2}; MMStar scores are taken from the ReVPT and Thyme papers~\cite{revpt,thyme}.
Published results retain the evaluation settings of their respective sources; the base model with a code interpreter and \ours{} use the protocol described in Section~\ref{sec:setup}.
Dashes indicate unreported scores.

\begin{table}[t]
\caption{Comparison with specialized visual tool-use models on perception benchmarks. Accuracy is reported in $\%$.}
\label{tab:specialized_tool_use}
\centering
\footnotesize
\setlength{\tabcolsep}{3pt}
\begin{tabular*}{\columnwidth}{@{\extracolsep{\fill}}lccccc@{}}
\toprule
\multirow{2}{*}{\textbf{Model}} & \multirow{2}{*}{\textbf{V*}} & \multicolumn{2}{c}{\textbf{HR-Bench}} & \multirow{2}{*}{\textbf{MMStar}} & \multirow{2}{*}{\textbf{TreeBench}} \\
\cmidrule(lr){3-4}
 & & \textbf{4K} & \textbf{8K} & & \\
\midrule
DeepEyes-7B~\cite{deepeyes} & 85.6 & 75.1 & 72.6 & --- & 37.5 \\
DeepEyesV2-7B~\cite{deepeyesv2} & 81.8 & 77.9 & 73.8 & --- & 42.5 \\
Pixel-Reasoner-7B~\cite{pixel_reasoner} & 84.3 & 74.0 & 66.9 & --- & 39.0 \\
ReVPT-3B~\cite{revpt} & --- & --- & --- & 53.9 & --- \\
ReVPT-7B~\cite{revpt} & --- & --- & --- & 61.1 & --- \\
Thyme-7B~\cite{thyme} & 82.2 & 77.0 & 72.0 & 65.9 & --- \\
\midrule
Baseline~\cite{qwen36_35b_a3b} & 90.6 & 80.9 & 74.9 & 78.7 & 56.5 \\
\ours{} (Ours) & 98.4 & 89.1 & 85.3 & 79.1 & 68.6 \\
\bottomrule
\end{tabular*}
\end{table}

\subsection{The Effectiveness of Coordinate Injections}
\label{sec:ablation_ci}
Table~\ref{tab:ci_ablation} evaluates inference-time Coordinate Injection on Gemini-3.5-Flash without updating its weights. Adding a code interpreter improves accuracy on several benchmarks. Applying CI further improves accuracy across all four evaluated benchmarks; on RealWorldCounting, accuracy increases from 37.4\% with uncorrected coordinates to 45.8\% with CI. These results support the value of correcting spatial operations during tool use and motivate the coordinate-corrected trajectories and spatial supervision used in Level~2 (Sections~\ref{sec:ci} and~\ref{sec:rl}).

\subsection{The Effectiveness of Grounding-Aware Process Reward}
\label{sec:ablation_reward}
We examine whether grounding-aware supervision improves spatial fidelity and final-answer accuracy during Level-2 RL after the Coordinate Injection cold-start.
The two variants share the same CI-based SFT initialization, tool interface, and optimizer: \emph{outcome-only} optimizes final-answer correctness ($\Racc$), whereas our \emph{full} reward additionally incorporates per-turn spatial fidelity ($\Rground$, Equation~\ref{eq:reward_total}).

Table~\ref{tab:grounding_curve} tracks grounding fidelity during training. Both variants start at $\bar{G}=0.55$. At step 200, the full-reward variant reaches $0.81$, compared with $0.63$ for outcome-only RL.
Although both series fluctuate, the full-reward variant maintains higher fidelity at every reported checkpoint after initialization, supporting the use of intermediate grounding supervision during agentic optimization.

Table~\ref{tab:reward_ablation} reports accuracy on RealWorldQA, BLINK, TreeBench, and RealWorldCounting.
The full reward improves accuracy over outcome-only RL on all four benchmarks, by 2.0, 3.0, 5.1, and 8.4 percentage points, respectively.

\begin{table}[t]
\caption{Effect of the grounding-aware process reward at Level~2, on the Qwen3.6-Flash (35A3) backbone.
Accuracy ($\%$); best per column in \textbf{bold}.}
\label{tab:reward_ablation}
\centering
\resizebox{\columnwidth}{!}{
\setlength{\tabcolsep}{4pt}
\renewcommand{\arraystretch}{1.15}
\begin{tabular}{@{}lcccc@{}}
\toprule
\textbf{Level-2 reward} & \textbf{RealWorldQA} & \textbf{BLINK} & \textbf{TreeBench} & \textbf{\shortstack{RealWorld\\Counting}} \\
\midrule
Outcome-only ($\Racc$) & 85.1 & 73.4 & 63.5 & 45.5 \\
\addlinespace[2pt]
$+$ Process reward (\textbf{Ours}) & \textbf{87.1} & \textbf{76.4} & \textbf{68.6} & \textbf{53.9} \\
\bottomrule
\end{tabular}
}
\end{table}

%% file: sec/6_conclusion.tex
\section{Conclusion}
\label{sec:conclusion}

We presented \ours{}, a framework for improving visual perception in MLLMs during the post-training stage.
By converting the model's perceptual errors into verifiable QA and combining CI-corrected trajectories with grounding-aware process reward, the framework provides training signals for both single-turn perception and multi-turn spatial grounding. The process-reward ablation shows improved grounding fidelity during training and final-answer accuracy gains across all four evaluated benchmarks.
Our two-level approach, Perception RL for single-turn QA and Perception Agentic RL for multi-turn tool use, improves aggregate performance across eight perception benchmarks, demonstrating that visual perception can be effectively optimized in post-training.
Future directions include extending to video perception with temporal consistency constraints, learning the grounding model jointly with the VLM, and applying CI to other agentic domains such as GUI interaction.